\documentclass[runningheads]{llncs}
\usepackage{graphicx}

\usepackage{tikz}
\usepackage{comment}
\usepackage{amsmath,amssymb} 
\usepackage{color}
\usepackage{orcidlink}
\usepackage{mciteplus}

\usepackage[accsupp]{axessibility}  

\begin{document}
\pagestyle{headings}
\mainmatter
\def\ECCVSubNumber{104}  

\title{End to End Face Reconstruction via Differentiable PnP} 

	\titlerunning{End to End Face Reconstruction via Differentiable PnP}
	%
	\author{Yiren Lu\inst{1}\thanks{Work done during an internship in Tencent. \\ Two authors contributed equally to this work.}\orcidlink{0000-0002-5411-0411}  \and
		Huawei Wei\inst{2}\orcidlink{0000-0001-9941-6636}
		}
	\authorrunning{Y. Lu and H. Wei}
	%
	\institute{State University of New York at Buffalo \\
		\email{yirenlu@buffalo.edu}\\ \and
		Tencent\\
		\email{huaweiwei@tencent.com}}
\maketitle
\begin{abstract}

This is a challenge report of the ECCV 2022 WCPA Challenge,
Face Reconstruction Track. Inside this report is a brief explanation
of how we accomplish this challenge.
We design a two-branch network to accomplish this task, whose roles are
Face Reconstruction and Face Landmark Detection. The former outputs canonical 3D face coordinates. 
The latter outputs pixel coordinates, 
i.e. 2D mapping of 3D coordinates with head pose and perspective projection.
In addition, we utilize a differentiable PnP (Perspective-n-Points) layer to finetune the outputs of the two branch.
Our method achieves very competitive quantitative results on the MVP-Human dataset and wins a $3^{rd}$ prize in the challenge.

\end{abstract}

\section{Introduction}
Face reconstruction from a single image has a wide range of 
applications in entertainment and CG fields,
such as make-up in AR and face animation.
The current methods \cite{Bai_2021_CVPR, https://doi.org/10.48550/arxiv.2012.04012, https://doi.org/10.48550/arxiv.1803.07835, guo2020towards} generally use a simple orthogonal projection in face reconstruction.
The premise of this assumption is that the face is far enough away from the camera.
When the distance is very close, the face will suffer from severe perspective distortion.
In this case, the reconstructed face has a large error with the real face.
To solve this problem, perspective projection must be adopted as the camera model.
In this paper, we propose a method to reconstruct 3D face under perspective projection.
We develop a two-branch network to regress 3D facial canonical mesh and 2D face landmarks simultaneously.
For 3D branch, we adopt a compact PCA representation \cite{guo2020towards} to replace the dense 3d points. 
It makes the network learn easier. For 2D branch, we take uncertainty learning into account.
Specifically, Gaussion Negative Log Loss \cite{NIPS2017_2650d608} is introduced as our loss function. 
It can enhance the robustness of the model, especially for faces with large poses and extreme expressions.
In addition, a differentiable PnP (Perspective-n-Points) layer \cite{chen2022epro} is leveraged to finetune our network after the two branches converge.
It brings interaction of the two branches, so that the two branches can be jointly trained and promote each other. Our method achieves very competitive results on the MVP-Human dataset.

\begin{figure}
	\centering
	\includegraphics[height=6.5cm]{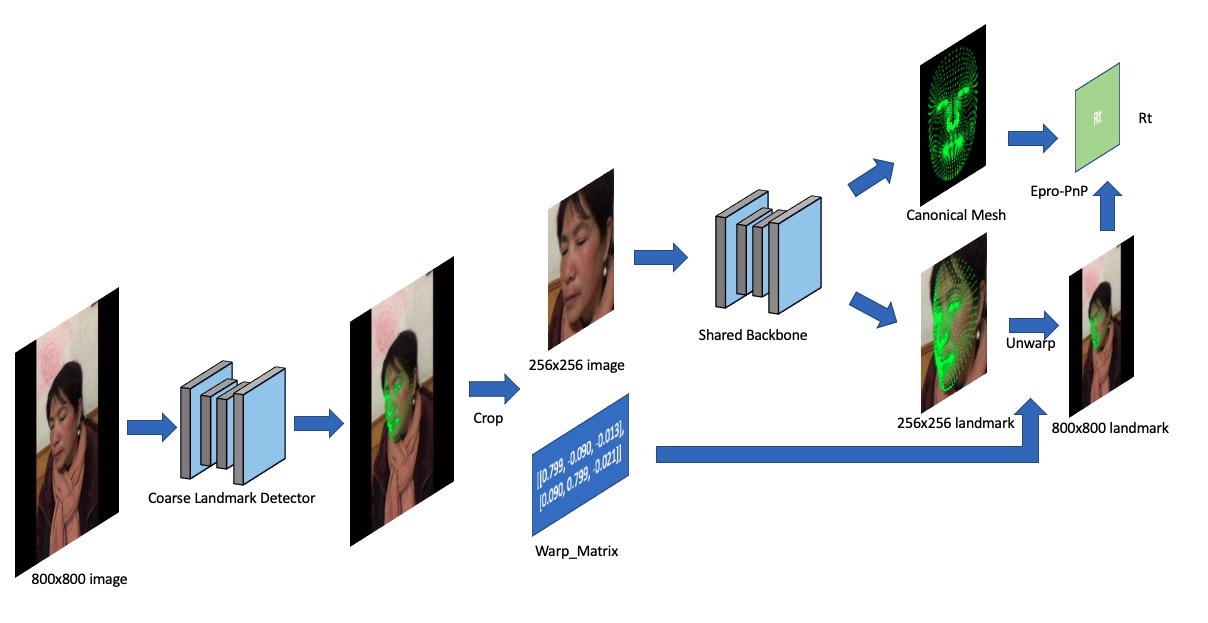}
	\caption{This figure shows the overall pipeline of our method. First, the 800$\times$800 image will go through a Coarse  Landmark Detector to get coarse landmarks so that we can utilize them to crop the face out. Then the cropped 256$\times$256 image will be fed into a shared backbone with different regressors. After that, a canonical mesh and a 256$\times$256 landmark will be output. Finally, we use the unwarped landmark and the canonical mesh to find out the Rotation and Translation through PnP.}
	\label{fig:pipeline}
\end{figure}

\section{Methodology}
We decouple the problem into two parts: head pose estimation and canonical
mesh prediction.
For head pose estimation, 
the commonly used methods are generally divided into two categories.
The first category is to directly regress the 6DoF pose from the image.
The second is to utilize PnP (Perspective-n-Points) to figure out the 6DoF pose with 
the correspondence between 2D and 3D coordinates.
We find it's hard for the network to directly regress the 6DoF well, especially for the translation of the depth axis.
So we choose the second category of method as our framework.
Our strategy is to regress canonical 3D mesh and 2D facial landmarks separately. To this end, we develop a two-branch
network, which respectively outputs canonical 3D face coordinates and the landmarks, i.e. the perspective projection of 3D coordinates.
When both branches converge, we use a differentiable PnP layer to jointly optimize the two branches.
Its purpose is to make the 6DoF pose calculated from 2D and 3D coordinates as close to the groundtruth pose as possible. 
In inference phase, we directly use PnP to compute 6DoF pose from the output of two branches.
An overview of our pipeline is showed in Fig. \ref{fig:pipeline}.

\subsection{Data Preparation}

The original image size is 800x800, training with images of this size is very inefficient. 
Hence, it is necessary to crop the image to an appropriate size.
To achieve this purpose, we train a network that predicts the 2D face landmarks
on 800×800 image. We utilize these landmarks to frontalize the face (rotate the face to a status that the roll is 0), and crop the face out. 
After this process, we obtain a 256×256 aligned
image and a warp matrix for each image, with which the predicted 256×256 landmarks
can be transformed back to the corresponding positions on the original
800×800 image.

\subsection{Facial Landmark Detection}
Generally, we consider Facial Landmark Detection as a regression task to find out the corresponding position of each landmark on the image. We also follow this strategy, but with some small modifications.
\subsubsection{Probabilistic Landmark Regression \cite{wood2022dense}}
Instead of regressing the coordinates $\mathbf{x}$ and $\mathbf{y}$ of each landmark point. 
We introduce Gaussion Negative Log Loss(GNLL) to regress the uncertainty of each point along with the coordinates:

\begin{equation}
L_{gnll} = \sum_{i=1}^{|Lmk|} \left(\log(\sigma_i^2)+ \frac{||\mu_i - \hat{\mu_i}||^2}{2\sigma_i^2}\right).
\label{1}
\end{equation}
In which $Lmk$ represents the set of all 1220 landmarks, and $Lmk_i =(\mu_i, \sigma_i)$. In $Lmk_i$, $\mu_i = (x_i, y_i)$ represents the ground truth coordinates of each landmark, and $\sigma_i$ indicates the uncertainty or invisibility of each landmark. We find that, with the help of GNLL, the predicted landmarks will be more accurate than training using regular MSE loss.

\subsubsection{CoordConv\cite{Liu2018AnIF}}

We apply CoordConv in our landmark
detection branch, which plays a position encoding role. With this technique, the input of the landmark
detection model would be 5 channels (RGBXY) instead of 3 channels (RGB). 
We find CoordConv helps to improve the accuracy of the landmark regression.	

\subsection{3D Face Reconstruction}

In this branch, we need to predict the coordinates of each point from the canonical
mesh. The most direct way is to roughly regress the coordinates. However,
we find this strategy not that ideal. Our strategy is to convert the mesh into a compact PCA representation.
It is verified the network training is easier and higher accuracy can be obtained with this strategy.

\subsubsection{Construct PCA Space}
We manually select a face with neutral expression from each of the 200 persons in the training set. 
In addition, to enhance the representation ability of PCA, 51 ARKit\footnote[1]{https://developer.apple.com/documentation/arkit/arfaceanchor/blendshapelocation} blendshape meshes are involved (the TongueOut blendshape is excluded).
We use these 251 face mesh to fit a PCA model, whose dimension of components is eventually 250.
Then the PCA model is utilized to fit the coefficient for each training sample.

\subsubsection{Loss Function}
We use Vertex Distance Cost (VDC) and Weighted Parameter Distance Cost (WPDC) proposed by 3DDFA V2 \cite{3ddfa_cleardusk, guo2020towards} as our loss function. And the algorithms are showed below, 

VDC loss:
\begin{equation}
	L_{vdc} = ||X_{3d}^{pred} - X_{3d}^{gt}||^2,
\end{equation}
where $X_{3d}^{pred}$ represents the 3d mesh predicted by network and $X_{3d}^{gt}$ represents the ground truth 3d mesh.

WPDC loss:
\begin{equation}
	L_{wpdc} = ||W \cdot (PCA^{pred} - PCA^{gt})||^2  ,
\end{equation}
where $PCA^{pred}$ represents the predicted PCA coefficients and $PCA^{gt}$ presents the ground truth ones, and $W$ represents the weight of each coefficient.

\subsection{PnPLoss}

In order to make the two branches have stronger interaction, we try to jointly optimize the two branches by a
differentiable PnP layer. Specifically, we utilize Epro-PnP \cite{chen2022epro} proposed recently on CVPR2022 to achieve this. Epro-PnP 
interprets pose as a probability distribution and replace the argmin
function with a softargmin in the optimization procedure. Hence, it is a differentiable operation and can be
plugged into neural network training. The inputs of the pnp layer is 2D landmarks and 3D mesh coordinates, then the pose
will be output. By making the pose as close to the groundtruth pose as possible, the coordinates locations of 2D/3D branch will be finetuned to 
a more accurate status. The expression of PnPLoss is showed below, 
\begin{equation}
P^{pred} = EproPnP(X_{2d}^{pred}, X_{3d}^{pred}) \label{2},
\end{equation}
\begin{equation}
	L_{pnp} = ||P^{pred}  X_{3d}^{gt} - P^{gt}  X_{3d}^{gt}||_2 
	+ ||P^{pred}  X_{3d}^{pred} - P^{gt}  X_{3d}^{gt}||_2\label{3},
\end{equation}
where $X_{2d}^{pred}$, $X_{3d}^{pred}$ represents the 3d mesh, 2d landmark predicted by network and $P^{pred}$ stands for the predicted pose output by Epro-PnP.

~\\
\noindent
Finally, the total loss function should look like this:
\begin{equation}
	L_{total} = \lambda_1 L_{gnll} + \lambda_2 L_{vdc}+ \lambda_3 L_{wpdc} + \lambda_4 L_{pnp}.
\end{equation}

\noindent
One thing need to be informed is that we do not train from scratch with this pnp layer.
We first train the network for several epoches without it. Then we use Epro-PnP
layer to finetune the training.

\subsection{Inference Phase}
In inference phase, we utilize opencv to implement the PnP module. Before start to solve PnP,
we first need to unwarp the landmarks predicted on 256×256 to 800×800 by
multipling the inverse matrix of the warp matrix.

\section{Experiments}
\subsection{Experimental Details}
We use the aligned 256x256 images as training samples and add color jitter and flip augmentation
in training process. Resnet50 \cite{he2016deep} (pretrained on ImageNet \cite{5206848})is utilized as our backbone. The regressor layer of 2D/3D branch is 
a simple FC layer. In training stage, we use Adam optimizer \cite{Kingma2015AdamAM} and 1e-4 as the learning rate. We first train for 10 epochs, 
and then finetune the model for 5 epochs with Epro-PnP. Besides, the $\lambda$'s in the total loss function are 0.01, 20, 10, 2 respectively.

\subsection{Head Pose Estimation}
In this experiment section we compare the performance of several methods to estimate the headpose Rt.  We choose $MAE_r$, $MAE_t$, $ADD$ as our evaluation metrics, where $MAE_r$ is the mean absolute error of the 3 components of eular angle, yaw, pitch and roll; $MAE_t$ is the mean absolute error of the 3 components of translation, x, y and z; $ADD$ is the second term of challenge loss, which is shown in Eq.\eqref{4}

\begin{equation}
	ADD = ||P^{gt}  X_{3d}^{gt} - P^{pred} X_{3d}^{gt}||_2 \label{4}.
\end{equation}

\subsubsection{Direct 6DoF with CoordConv}
This is our baseline method. As we are trying to regress Rt which corresponds to the 800$\times$800 image on a 256$\times$256 image, we need to give it some extra information. So we add a 2 channel coordmap XY to indicate where each pixel in the 256$\times$256 image locates on the original 800$\times$800 image. And then feed the 5 channel input to a ResNet50 backbone and a linear regressor. The output should be 9 numbers, 6 of them for Rotation and 3 for Translation. We supervise the training process with Geodesic Loss \cite{hempel20226d} and MSE Loss for Rt. From Table 1, we can see that though it cannot compete with PerspNet \cite{kao2022single}, it is much better than Direct 6DoF without coordmap.

\subsubsection{MSELoss \& PnP} In this experiment we try to use landmark detection together with PnP to calculate Rt, the choice of backbone and regressor is the same as the previous experiment. The input of network is simply 3 channel RGB image and we utilize only landmark MSE loss to supervise the training process. We can see that it outperforms PerspNet and result in a 9.79 ADD loss.

\subsubsection{GNLL \& PnP} In this experiment, we increase the input of the network from 3 channels to 5 channels, we add a 2 channel coordmap XY(the range of coordmap is [0, 255] and normalized into [-1, 1]) and concatenate them with RGB channels. Besides, we utilize GNLL to supervise the training process instead of landmark MSE loss. This experiment reaches a 9.66 ADD loss, which is slightly better than the previous one.

\begin{table}
	\begin{center}
		\caption{Comparison with the performance of different methods for 6DoF Head Pose Estimation on validation set. The experiment result of Direct 6DoF and PerspNet is from the PerspNet \cite{kao2022single} paper.}
		\label{table:headings}
		\begin{tabular}{llll}
			\hline\noalign{\smallskip}
			Method & $MAE_r$ & $MAE_t$ & ADD \\
			\noalign{\smallskip}
			\hline
			\noalign{\smallskip}
			Direct 6DoF  & 1.87 & 9.06 & 21.39\\
			Direct 6DoF(CoordConv)  &1.11& 5.58 & 13.42 \\
			PerspNet & 0.99 & 4.18  & 10.01\\
			MSELoss \& PnP & 0.89 & 4.02 & 9.79 \\
			\textbf{GNLL \& PnP}  & \textbf{0.82} & \textbf{3.95} & \textbf{9.66}\\
			
			\hline
		\end{tabular}
	\end{center}
\end{table}

\subsection{Face Reconsturction}
\subsubsection{Direct 3D points}
In this experiment we try to directly regress each 3d coordinate of the canonical mesh. Points MSE Loss is used to supervise the training process. From Table 2, we can see that the result is not that satisfying, it only reaches a 1.82 mean error.

\subsubsection{Regress PCA Coefficient}
Instead of regressing 3D coordinates, we try to regress 250 PCA coefficients, which should be easier for the network to learn. Also, backbone and regressor remains the same as in the Direct 3D experiment. Just as the way we think, PCA results in a 1.68mm mean error and has performed better than PerspNet.

\begin{table}
	\begin{center}
		\caption{Comparison with the performance of different methods for Face Reconstruction on validation set. The experiment result of PerspNet is from the PerspNet paper.}
		\label{table:headings}
		\begin{tabular}{llll}
			\hline\noalign{\smallskip}
			Method & Medium(mm) \quad \quad & Mean(mm) \\
			\noalign{\smallskip}
			\hline
			\noalign{\smallskip}
			PerspNet & 1.72 & 1.76\\
			Direct 3D  \quad \quad&1.79 & 1.82\\
			\textbf{PCA} & \textbf{1.63} & \textbf{1.68} \\
			\hline
		\end{tabular}
	\end{center}
\end{table}

\subsection{Finetuning with PnP Layer} In this experiment, we introduce the newly proposed Epro-PnP. Instead of training from scratch, we choose to finetune the network based on the previous experiment. After finetuning, we  can find from Table 3 that $MAE_t$ and ADD loss reduces obviously. 
We attribute the effectiveness to that the pnp layer can enhance the
interaction of the 2D/3D branches.
It makes the two branches not only for the pursuit of lower coordinates regression error, 
but to make the pose estimation more accurate.

\begin{table}
	\begin{center}
		\caption{Comparison with the performance for 6DoF Head Pose Estimation with and without Epro-PnP on validation set.}
		\label{table:headings}
		\begin{tabular}{llll}
			\hline\noalign{\smallskip}
			Method & $MAE_r$ & $MAE_t$ & ADD \\
			\noalign{\smallskip}
			\hline
			\noalign{\smallskip}
			GNLL \& PnP  & 0.82 & 3.95 & 9.66\\
			\textbf{GNLL \& PnP \& Epro-PnP} \quad \quad& \textbf{0.81}  & \textbf{3.86} &\textbf{9.48} \\
			
			\hline
		\end{tabular}
	\end{center}
\end{table}

\subsection{Qualitative Results}
\begin{figure}
	
	\centering
	\rotatebox{90}{\scriptsize{~~~~~GT Result~~~~~~~~Our Result~~~~~~~~~~~Input}}
	\includegraphics[width=4.5in]{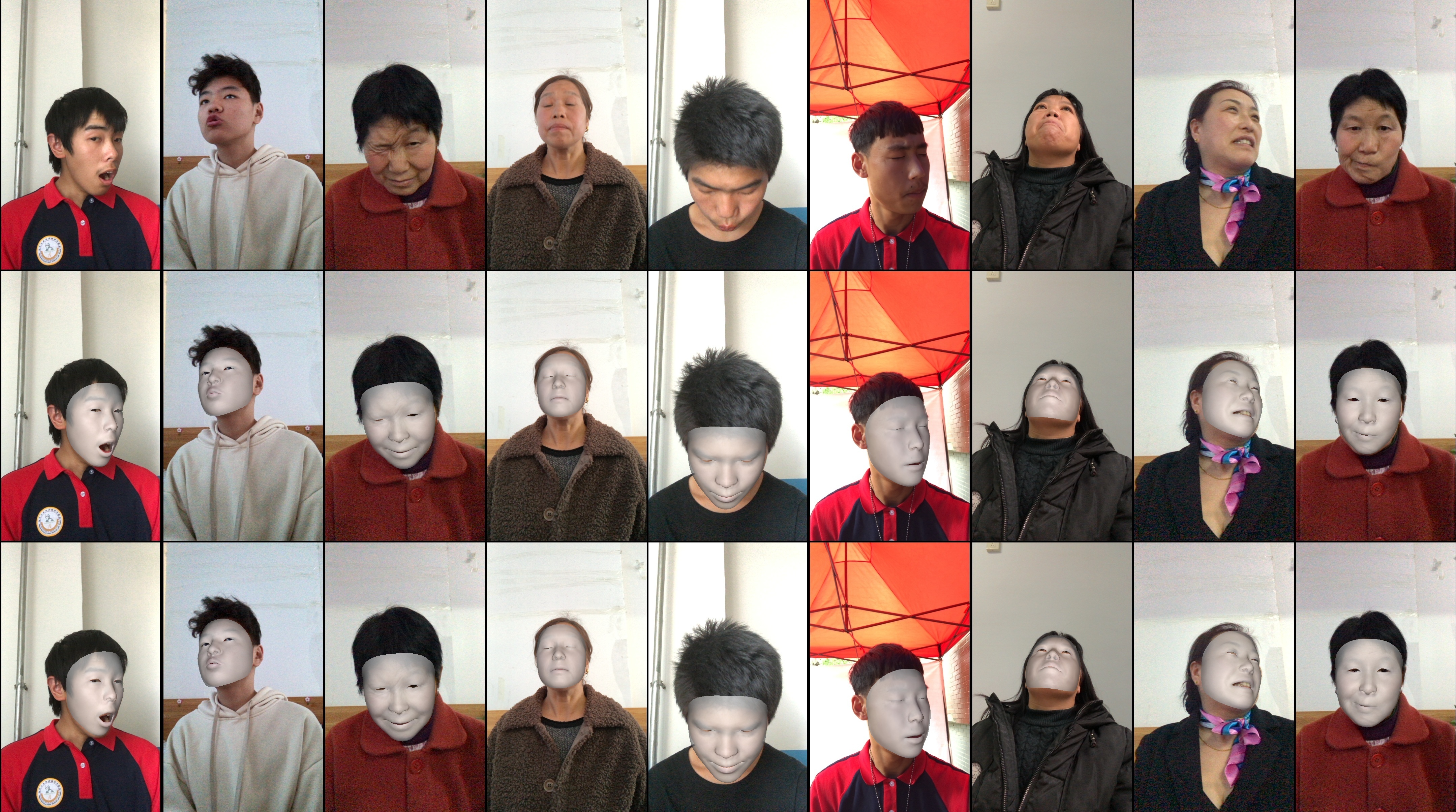}
	\caption{Visualization comparison between our method and the ground truth obtained by the structured light sensor of iphone11. We can find that in some extreme cases such as the first column and the fifth column, our method outperforms the ground truth result, which means we have learned the essence of this task.}
	\label{fig:qualititive result}
\end{figure}

We show some qualitative results in Fig. \ref{fig:qualititive result}. 
It can be observed that our method can accurately reconstruct faces even in large poses and extreme expressions.
In some cases, our method even outperforms the ground truth.

\section{Tricks}

\subsection{Flip \& Merge}
When doing inference on test set, we flip the input image and let the flipped and unflipped go through the network at the same time. After getting two results, we merge them and calculate an average as the final result. One thing worth noting is that this strategy only works in the face reconstruction branch in our test, in the landmark detection branch it is ineffective. We think this is due to the face reconstruction task with PCA is to some extent similar to a classification problem. 

\section{Conclusions}
In this competition, we design a two-branch network to solve the 3D face reconstruction task.
Meanwhile, we propose to utilize a differentiable PnP layer to jointly optimize the two branches.
Finally, our method achieves a competitive result and scores 33.84 on leaderboard with a $3^{rd}$ rank.

%
%
\bibliographystyle{splncs04}
\bibliography{egbib}
\end{document}